\documentclass[10pt,journal,compsoc]{IEEEtran}
\usepackage{rotating, graphicx}
\usepackage{cite}
\usepackage{amsmath,amssymb,amsfonts}
\usepackage{algorithm}
\usepackage{algorithmic}
\usepackage{graphicx}
\usepackage{textcomp}
\usepackage{algorithmic}
\usepackage{subfigure}
\usepackage{graphicx}
\usepackage{amssymb}
\usepackage{url}
\usepackage{enumerate}

\usepackage{rotating}

\begin{document}
%
\title{{\bf O}ne-{\bf C}lass {\bf C}lassification by {\bf E}nsembles of {\bf R}egression models - a detailed study  }
%
%
%
%

\author{Amir Ahmad* and Srikanth Bezawada
\IEEEcompsocitemizethanks{\IEEEcompsocthanksitem Amir Ahmad and Srikanth Bezawada are with the College of Information Technology, United Arab Emirates University, Al-Ain, UAE
\protect\\
* Corresponding author, E-mail: amirahmad@uaeu.ac.ae 
}
\thanks{}}

\IEEEtitleabstractindextext{%
\begin{abstract}
One-class classification (OCC) deals with the classification problem in which the training data has data points belonging only to target class. 
In this paper, we study a one-class classification algorithm, {\bf O}ne-{\bf C}lass {\bf C}lassification by {\bf E}nsembles of {\bf R}egression models (OCCER), that uses regression methods to address OCC problems.  The OCCER  coverts an OCC problem into many regression problems in the original feature space so that each feature of the original feature space is used as the target variable in one of the regression problems. Other features are used as the variables on which the dependent variable depends. The errors of  regression of a data point by all the regression models are used to compute the outlier score of the data point.  An extensive comparison of the OCCER algorithm with state-of-the-art OCC algorithms on several datasets was conducted to show the effectiveness of the this approach. We also demonstrate that the OCCER algorithm can work well with the latent feature space created by autoencoders for image datasets. The implementation of OCCER is available at https://github.com/srikanthBezawada/OCCER.
\end{abstract}

}

\maketitle

\IEEEdisplaynontitleabstractindextext

%
\IEEEpeerreviewmaketitle

\IEEEraisesectionheading{\section{Introduction}\label{sec:introduction}}
The one-class classification (OCC) problem is a special class of classification problems in which only the data points of one class (the target set) are available \cite{Tax2001}.  The task in one-class classification is to
make a model of a target set of data points and to predict if a testing data point is similar to the target set. Point which are not similar to the target set are called outlier. OCC algorithms have applications in various domains \cite{OCCreview2} including
anomaly detection, fraud detection, machine fault detection, spam detection, etc. \cite{OCCreview2}.

The OCC problem is generally considered to be more difficult problem than the two-class classification problem as the training data has only data points belonging to one class \cite{Tax2001,occreview1,OCCreview2}, and traditional classifiers need training data from more than one class to learn decision boundaries. Therefore, standard classifiers cannot be applied directly to OCC problems. Various algorithms have been proposed to address OCC problems \cite{Tax2001,occreview1,OCCreview2}. 

There are two main approaches to handle OCC problems \cite{occreview1,OCCreview2}. In the first approach, artificial data points for the non-target class (outlier) are generated and combined with the target data points and then a binary classifier is  trained on this new data. In the second approach, target data points are used to create the OCC models \cite{DESIR20133490}.  

Gaussian models \cite{occreview1}, reconstruction-based methods \cite{Tax2001,occreview1,OCCreview2}, nearest neighbours \cite{Tax2001,jknn}, support vector machines \cite{m._j._tax_support_1999,scholkopf_support_2000} clustering based methods \cite{Tax2001} and convex hull \cite{convexhull} are some examples of the second approach.

Ensembles of accurate and diverse models generally perform better than individual members of ensembles \cite{Dietterich}. Ensembles of classification models have been developed to improve the performance of one-class classification models \cite{IsolationForests,jknn,autoensembles,clusterensembles}.

Regression analysis is a well-established research area \cite{Bishop:2006:PRM:1162264}. In a regression problem, the relationship between a dependent variable and other independent variables is investigated. Various regression methods have been developed \cite{Bishop:2006:PRM:1162264}.

Nato et al. \cite{Nota} propose OCCER that uses ensembles of regression models to address anomaly detection problem. They test OCCER only on UCI datasets \cite{Dua:2019}. In this paper, we carried out an extensive study of OCCER on various kinds of datasets. We also studied the performance of OCCER on image datasets \cite{mnist} by creating latent features using autoencoders. The effect of ensemble size on the performance of OCCER was also investigated.

The paper is organised as follows. Section 2 discusses about related work. The OCCER algorithm is presented in Section 3. Section 4 presents experiments and discussion. Section 5 discusses the conclusion and suggests future developments. 

\section{Literature Survey}
As previously discussed there are two types of OCC algorithms, and  
OCCER belongs to the second type, which will be discussed in this section.

Generative methods are useful for OCC as the target class may directly be modelled from the available training target data points. Density-based methods, such as Gaussian, kernel density estimators, Parzen windows and mixture models  are widely used for OCC problems \cite{occreview3,occreview1},. Density-based methods estimate the probability density function of the underlying distribution of the training target data points. Then, these methods determine if a new data point comes from the same distribution.  The selection of appropriate models and  large-scale training data are the problems of this approach. 

Nearest neighbour-based (NN-based) approaches are other widely used methods to address OCC problems\cite{Tax2001,occreview1,jknn}. This approach assumes that an outlier point will be far away from neighbour target points as compared to a target point from other neighbour target points \cite{Tax2001,jknn}.

The local outlier factor (LOF) method is a density-based scheme for OCC \cite{LOF}, in which a LOF is computed for each data point by taking the ratios of the local density of the point and the local densities of its neighbours. An outlier point has a large LOF score.

Tax and Duin \cite{m._j._tax_support_1999} propose the support vector domain description method for OCC. The method finds a hyper-sphere with a minimum volume around the target class data such that it encloses almost all the points in the target class dataset. Scholkopf et al. \cite{scholkopf_support_2000} propose the use of support vector machines  for one class classification. A hyperplane is constructed such that it separates all the data points from the origin and the hyperplane's distance from the origin is maximised.

In reconstruction-based methods \cite{Tax2001,occreview1,OCCreview2}, a model like autoencoder is trained on the given target class data. The reconstruction error which depends on a testing data point and  the system output is used to define the outlier score. An outlier point is likely to have more reconstruction error. 

Clustering-based approaches use a clustering method, like k-means clustering to create clusters \cite{Tax2001}. The distance between a data point and its nearest cluster centre is used as the outlier score. The number of clusters and cluster initialization are the problem of k-means type clustering algorithms.

Arashloo and  Kittler  \cite{kernelregression} present a nonlinear one-class classifier formulated as the Rayleigh quotient criterion
optimisation that projects the target class points to a single point in a new feature space, the distance between the projection of a testing point to that point is the outlier score of the testing point. Leng et al. \cite{Ridgeregression} use a similar approach,but uses extreme learning machines for the projection.     

 Ensembles have also been developed for the OCC problems. There are two approaches for creating ensembles. In the first approach, one OCC algorithm is employed and a randomisation process is used to create diverse OCC models. Lazarevic and Kumar \cite{featurebagging} propose the creation of multiple datasets by using feature bagging. The LOF algorithm is then used on these multiple datasets, hence multiple OCC models are created. The outputs of these models are combined to get the final output. Khan and Ahmad \cite{jknn} use random projection to create multiple datasets. A NN-based OCC algorithm is applied to these multiple datasets. Zimek et al. \cite{Zimekper} introduce noise to the dataset to create multiple datasets. Experiments with different OCC algorithms show the effectiveness of the proposed approach. Chen et al. \cite{autoensembles} use  randomly connected autoencoders to create ensembles of autoencoders. These ensembles outperformed other state-of-the art OCC methods. Khan and  Taati \cite{Khanensembles} train different autoencoders using different features to create ensembles of autoencoders. They show that ensembles perform better than single autoencoders. Isolation forests consist of many decision trees \cite{IsolationForests}. These trees are created by using random partitioning. The authors argue that anomalies are susceptible to isolation and therefore have short path lengths. The method has produced excellent results on various datasets. Nato et al. \cite {Nota} propose OCCER that uses an ensemble of regression  models to produce outlier score. In the next section, we will discuss this model in detail.

\section{OCCER Algorithm } In this section, we discuss OCCER \cite{Nota}, an OCC algorithm, that uses an ensemble of regression models to compute the outlier score of a data point . 
For an $m$-dimensional dataset, OCCER converts the OCC problem into $m$ regression problems. Each regression problem generates one regression model. Each regression model produces an outlier score to a data point. $m$ outlier scores are combined to produce an outlier score for the data point. Firstly, we will present the motivation of the OCCER algorithm.

$D$ is a training dataset consisting of $n$ data points. Each data point is defined by $m$ features $\{x_1, x_2,.., x_i,..x_m\}$, and all data points belong to the target class. 
The rules of the target class are defined by using the following expression
\begin{eqnarray}
\mbox{\small \{values of $x_1$\}$\land$ \{values of $x_2$\}...$\land$ \{values of $x_i$\}...$\land$ \{values of $x_m$\}}
\end{eqnarray}

Classifiers like decision trees can learn these rules if the dataset has more than one class, however, with one class dataset, this is not possible. OCCER algorithm uses the information that all data points belong to the target class  to covert a OCC problem into $m$ regression problems. 
The rules to represent the class are defined by $m$ features, and these features are related to each other. OCCER uses regression models to learn these relations. One of the $m$ features is taken as the dependent variable, and the other $m -1$ features are taken as independent variables, and a regression model is trained on it. The regression function is presented as follows:

\begin{equation}
x_i^{'} = f^{i}(x_1,x_2,..,x_{i-1},x_{i+1},..x_m)
\end{equation}

Since $m$ features exist, $m$ different regression models can be trained. For a given data point, each regression model predicts one of the feature (which is used as the output feature) value. The predicted values and actual values are used to compute the outlier score (O) of the data point. The average of the absolute difference between the actual values and the predicted value for all the $m$ features is taken as the outlier score of the point. 

\begin{equation}
O = \sum_{i=1}^{m}|x_i^{'} - x_i|/m
\end{equation}

It is expected that data points which do not belong to the target class have larger outlier scores. The relationships between the features for the outlier data points will be different from the  relationships between  the features for the data points that belong to the target class (for which regression models are trained). Regression models will be less accurate for an outlier data point. It will lead to larger outlier scores for points that do not belong to the target class. Nota et al. \cite{Nota} propose to use accuracy of the prediction in computing outlier score. However, in this paper, we are using the Eq. 1 as the outlier score.  

\begin{table}[!ht]
\centering
\caption{An example dataset.}
\label{tab:toy}
\begin{tabular}{|l|l|l|l|l|}
\hline
Data point &$x_1$  & $x_2$  & $x_3$ & $x_4$
\\
\hline
1 & 0.85 & 0.34 &0.87 & 0.45
\\2&0.67& 0.43 & 0.43&0.54
\\3 & 0.79 & 0.89 & 0.63&0.71
\\- & -& -& -&-
\\-& - & -& -&-
\\
\hline
\end{tabular}
\end{table}

We explain the mechanism of the OCCER algorithm using the example dataset presented in Table 1. The dataset has points belonging to a target class, and  has four features:  $x_1, x_2, x_3 and x_4$. Four regression models will be created from the dataset, and each regression model uses one of the features as the dependent variable and all other three features as independent variables. 
Note that different features may have different ranges, so it is important to bring them on the same scale. Data normalisation is a necessary preprocessing step of OCCER.

\begin{equation}
x_1^{'} = f^{1}(x_2,x_3,x_4)
\end{equation}

\begin{equation}
x_2^{'} = f^{2}(x_1,x_3,x_4)
\end{equation}

\begin{equation}
x_3^{'} = f^{3}(x_1,x_2,x_4)
\end{equation}

\begin{equation}
x_4^{'} = f^{4}(x_1,x_2,x_3)
\end{equation}

These functions compute the outlier score  for a given data point. For example, the outlier score of data point 1 can be computed using Eq. 3,

\begin{eqnarray*}
Outlier score = (|f^{1}(0.34,0.87,0.45)-0.85| + \\ |f^{2}(0.85,0.87,0.45)-0.34| + \\|f^{3}(0.85,0.34,0.45)-0.87| + \\|f^{4}(0.85,0.34,0.87)-0.45|)/4
\end{eqnarray*}

The OCEM algorithm is presented as Algorithm 1.
\begin{algorithm}[!ht]
		\caption{OCCER algorithm}
	\label{kmeansalgo}
		\begin{algorithmic}
		\STATE {\bf Input-} An $m$-dimensional training dataset \textit{T}.
		\STATE Begin
		\STATE Training phase
      \STATE Normalise data (compute z-score for each attribute value using the mean and standard deviation for each feature)
		\FOR{i=1...$m$} 
        \STATE Train a regression model, $f^{i}(x_1,x_2,..,x_{i-1},x_{i+1},..x_m)$, using one of the features $x_i$ as the dependent variable and all other features as independent variables. 
\ENDFOR
       \STATE Testing phase
      \STATE To compute the outlier score of a data point, normalise the data point using the steps as the training data points are normalised. 
		\STATE  $d$ = $\{d_1, d_2,--, d_i,--,d_m\}$ is the normalised data point.
		\STATE sum = 0.
		\FOR{i=1...$m$} 
        \STATE Compute the output by using the function                $f^{i}(d_1,d_2,..,d_{i-1},d_{i+1},..d_m)$ and find the  difference with $d_i$, 
        \STATE sum = sum + $|f^{i}(d_1,d_2,..,d_{i-1},d_{i+1},..d_m) - d_i|$.
\ENDFOR
        \STATE  Outlier score = sum/$m$.
		\STATE End 
\end{algorithmic}
\end{algorithm} 

\section{Experiments} We conducted experiments by using the scikit-learn python package (https://scikit-learn.org/stable/) and PyOD (a Python toolbox for scalable outlier detection) \cite{PyOD}. As OCCER can work with any regression method, experiments were carried out using the following: Ridge regression, Lasso regression, Elastic net regression \cite{Bishop:2006:PRM:1162264}  and Random Forests regression \cite{randomforests}. In this paper, we named  the OCCER with a regression method as OCCER(Regression method). 
 scikit-learn's regression APIs are used for these regression methods. Different standard OCC algorithms, Isolation Forests (IF), One-class SVM (OCSVM), LOF and autoencoders, were used for the comparative study. PyOD was used for these methods. The default parameters for these methods given in PyOD  were used in the experiments. $5\times 2$-fold cross-validation was used for the experiments. 
Stratified k-fold was implemented using scikit-learn to ensure that the folds were made by preserving the percentage of the samples for each class.
Only the target class points in the training data was used to train the OCC algorithms. Z-normalisation was used to normalise the data. As classification accuracy is not a correct performance matrix due to the highly-imbalanced testing data, we used the average area under the curve (AUC) for the receiver operating characteristics(ROC) curve as it is generally used to measure the performance of the OCC algorithms \cite{IsolationForests,autoensembles}. 

\subsection{Standard datasets and domain datasets}
 Various kinds of datasets were used in the experiments \cite{keeldata,plos,mobifall,dlr,Coventry,promisedata}, Some datasets are created  as imbalanced datasets \cite{keeldata,plos}. Information on these datasets is presented in Table 2. The domain datasets \cite{mobifall,dlr,Coventry,promisedata} belong to two different domains datasets: normal activity-fall activity datasets and software engineering-related datasets. The domain datasets \cite{mobifall,dlr,Coventry,promisedata} are naturally imbalanced datasets. Mobilfall data \cite{mobifall} was collected using  Samsung Galaxy S3 mobile
employing the integrated 3D accelerometer and gyroscope. The data has two classes normal activity and fall activity.  We used the data collected from 11 subjects who performed various normal and fall activities. We grouped the German Aerospace Centre (DLR) data \cite{dlr} into; normal activity and fall activity,  and only used the data
from the accelerometer and gyroscope. Only data from those subjects who performed both the activities were used. Coventry dataset (COV) \cite{Coventry} also has two classes normal activity and fall activity, and the complete information of these domain datasets is presented in detail in \cite{jknn}. We also carried out experiments on software engineering realted datasets \cite{promisedata}, the information of which is presented in Table 3.

The results (average AUC) for datasets presented in Table 2 are provided in Table 4, which suggest that out of 29 datasets, OCCER algorithms performed best for 18 datasets. and among the OCCER algorithms, the OCCER(RF) algorithm performed best.  OCCER(RF) algorithm performed best for 10 datasets. 
The results (average AUC)  for domain datasets (presented in Table 3) are provided in Table 5. The OCCER(RF) algorithm performed best for six datasets, whereas other non-OCC algorithms were best for four datasets, suggesting that OCCER(RF) was the best among the OCCER algorithms.

The Friedman post-hoc test with the Bergmann and
Hommel's correction \cite{Demsar,Garcia:Herrera:09} was used to compare the OCCER(RF) algorithm with four standard OCC algorithms for all the datasets presented in Tables 2 and 3 (39 datasets). In this test, the p-values for each pair of classifier are computed and then corrected by using the
Bergmann and Hommel's procedure. Two classifiers are significantly different, if the  $p$ value is less than 0.05. The results are presented in Table 6, which indicate that OCCER(RF) is statistically better than IF, OCSVM, and autoencoders, whereas no statistically significant difference exists between OCCER(RF) and LOF. However, LOF does not exhibit any statistically significant improvement on OCSVM and IF. This analysis reveals that the OCCER(RF) algorithm is the best among these algorithms.

\subsection{MNIST image datasets} In OCCER algorithms, we will have to train $m$ regression models. For some datasets, such as image datasets the number of features can be very large. We conducted experiments to study the performance of  OCCER algorithms in a new feature space. We used the MNIST dataset \cite{mnist} for this study, which contains images of 28$\times$28 greyscale handwritten digits. Autoencoders with five hidden layers (512, 128, 64, 128, 512) were used to create latent representation of images, these latent features were used  as new features. The outputs from the nodes of the middle layer were used as the new features. Hence, 64 features were created for each image. 
Datasets were created by selecting data points for a digit which act as target class data points and the data points for other digits were used as outlier data points. For example, 
mnist-0 data contains 2000 randomly selected images of digit 0 (target class), whereas 25 images of each of the other digits (225 images) are outlier data points. Ten datasets, 
one for each digit as the target class were created.  
A setup similar to that discussed in the previous section was used in the experiment. 
 We also used the autoencoders, which were utilised to create latent representations as OCC models for the experiments,  and we named them Autoencoder(I). The results (average AUC) are presented in Table 7, which suggest that OCCER algorithms performed best for seven out of ten datasets. Autoencoder(I) performed best for two datasets, whereas LOF performed best for one dataset. This is an interesting result as OCCER algorithms were trained on the latent representation of the images whereas Autoencoder(I) was trained on the image data. This indicates that OCCER algorithms can also be used for the datasets in a new feature space.
\subsection{Effect of the ensemble size} In the experiments, the size of the ensembles for OCCER is equal to the number of features. In other words, each feature is used as the output in one of the regression models. It is possible that some of the regression models may not be very accurate. Hence, the question arises whether  OCCER can  perform better with fewer number of accurate regression models. 
We conducted experiments to investigate this issue and selected the most accurate regression models to compute the outlier scores. The root-mean-square error for the training dataset was used to determine the accuracy of a regression model. We conducted experiments with the top 25\%, 50\%, and 75\% most accurate models of the total regression models. 

 The results for OCCER(ridge), OCCER(lasso), OCCER(elastic) and OCCER(RF) are presented in Table 7, which show that there is no consistent performance advantage for any ensemble size. For example, for OCCER(RF) using all the regression models generated the best results for six out of ten datasets. For example, for OCCER(RF), using all the regression models generated the best results for 6 out of 10 datasets. For two datasets, the top 75\% most accurate regression models gave the best results, whereas for the remaining two datasets, the top 50\% most accurate regression models gave the best performance. Similar results were observed with other datasets presented in Table 1. On the basis of our observations, we can suggest that it is better to use all the regression models in computing the outlier scores. However, it will be interesting to investigate the best ensemble size for OCCER.
\begin{table}[]
\scriptsize
\caption{Information on the datasets that were taken from  \cite{keeldata,plos}. The datasets presented before the separating line in the Table is taken from \cite{keeldata} whereas the datasets presented after the separating line is taken from \cite{plos}}.
\begin{tabular}{|l|l|l|l|}
\hline
Dataset                      & Features & Target class & Outlier class \\
\hline
pima                         & 8        & 500      & 268      \\
segment0                     & 19       & 1979     & 329      \\
winequality-red-4            & 11       & 1546     & 53       \\
winequality-red-8\_vs\_6     & 11       & 638      & 18       \\
winequality-white-3\_vs\_7   & 11       & 880      & 20       \\
winequality-red-8\_vs\_6-7   & 11       & 837      & 18       \\
winequality-white-3-9\_vs\_5 & 11       & 1457     & 25       \\
winequality-red-3\_vs\_5     & 11       & 681      & 10       \\
yeast1                       & 8        & 1055     & 429      \\
yeast3                       & 8        & 1321     & 163      \\
yeast-2\_vs\_4               & 8        & 463      & 51       \\
yeast-0-3-5-9\_vs\_7-8       & 8        & 456      & 50       \\
yeast-0-2-5-6\_vs\_3-7-8-9   & 8        & 905      & 99       \\
yeast-0-2-5-7-9\_vs\_3-6-8   & 8        & 905      & 99       \\
yeast-0-5-6-7-9\_vs\_4       & 8        & 477      & 51       \\
yeast-1-4-5-8\_vs\_7         & 8        & 663      & 30       \\
yeast-1-2-8-9\_vs\_7         & 8        & 917      & 30       \\
yeast4                       & 8        & 1433     & 51       \\
yeast5                       & 8        & 1440     & 44       \\
yeast6                       & 8        & 1449     & 35       \\
\hline  
aloi-unsupervised            & 27       & 48492    & 1508     \\
annthyroid-unsupervised      & 21       & 6666     & 250      \\
breast-cancer-unsupervised   & 30       & 357      & 10       \\
letter-unsupervised       & 32       & 1500     & 100      \\
satellite-unsupervised       & 36       & 5025     & 75       \\
shuttle-unsupervised         & 9        & 45586    & 878      \\
speech-unsupervised          & 400      & 3625     & 61       \\
pen-global-unsupervised      & 16       & 719      & 90       \\
pen-local-unsupervised       & 16       & 6714     & 10 \\   
\hline  
\end{tabular}
\end{table}

\begin{table}[]
\scriptsize
\caption{Information on the domain datasets.}
\begin{tabular}{|l|l|l|l|}
\hline
Dataset                      & Features & Target & Outlier \\
                             &          & class & class \\
\hline
MobiFall (MF) \cite{mobifall}   & 31 & 5430 & 488  \\
Coventry dataset (COV) \cite{Coventry}  & 31 &12392 & 908 \\
German Aerospace Centre (DLR) \cite{dlr} & 31 &26576  & 84 \\
\hline 
Software related datasets \cite{promisedata} & & & \\
\hline
class-level-kc1-(defective or not)          & 94 &  85 & 60   \\
kc2-Software defect prediction                                 & 21 & 415 & 105  \\
kc1-Software defect prediction                                  & 21 & 1783 & 326  \\
cm1-Software defect prediction                                  & 21 & 449  & 49   \\
datatrieve-Software defect prediction                          & 8  & 119  & 11   \\
pc1-Software defect prediction                                  & 21 & 1032  & 77  \\
class-level-kc1-defect-count-ranking                     &  94  &  137   & 8\\      
\hline
\end{tabular}
\end{table}
\onecolumn
\begin{table}{}
\caption{Average AUC of various OCC algorithms against the OCCER algorithm with different regression methods on various datasets  \cite{keeldata,plos} presented in Table 2. Bold numbers indicate the best performance.} 
	\begin{tabular}{|l|l|l|l|l|l|l|l|l|}
\hline
	Dataset                      & IF        & LOF         & OCSVM          & Autoencoder  & OCCER   & OCCER   & OCCER & OCCER
\\    &    &  & & &(ridge) &(lasso) &(elastic) &(RF) 
      \\
\hline
	pima                         & \textbf{0.73153} & 0.70921          & 0.70083              & 0.64848          & 0.69958          & 0.70826          & 0.70998          & 0.69287          \\
segment0                     & 0.47464          & 0.8156           & 0.29403              & 0.34288          & 0.74055          & 0.3188           & 0.31852          & \textbf{0.89558} \\
winequality-red-4            & 0.58476          & 0.65112          & 0.61545              & 0.60903          & 0.56284          & 0.60235          & 0.60405          & \textbf{0.67001} \\
winequality-red-8\_vs\_6     & 0.6676           & 0.59251          & 0.64706             & 0.68153          & 0.56468          & 0.68899          & \textbf{0.69732} & 0.67908          \\
winequality-white-3\_vs\_7   & 0.84964          & 0.86687          & 0.85383             & 0.85111          & 0.85955          & 0.84157          & 0.8397           & \textbf{0.91371} \\
	winequality-red-8\_vs\_6-7   & 0.59562          & 0.54201          & 0.5876               & 0.6172           & 0.51581          & 0.61935          & \textbf{0.62417} & 0.60288          \\
	winequality-white-3-9\_vs\_5 & 0.80361          & 0.79305          & 0.79046              & 0.79714          & 0.72918          & 0.78964          & 0.79132          & \textbf{0.82647} \\
	winequality-red-3\_vs\_5     & 0.79092          & 0.81772          & 0.81567             & 0.79785          & 0.79632          & 0.7984           & 0.80191          & \textbf{0.82155} \\
	yeast1                       & 0.54367          & \textbf{0.61541} & 0.54807              & 0.53404          & 0.52579          & 0.52303          & 0.52322          & 0.58466          \\
	yeast3                       & 0.67375          & \textbf{0.80712} & 0.72526              & 0.72837          & 0.75759          & 0.65604          & 0.66245          & 0.78351          \\
	yeast-2\_vs\_4               & 0.87973          & 0.89002          & 0.89062             & \textbf{0.89893} & 0.83079          & 0.89442          & 0.8944           & 0.82599          \\
	yeast-0-3-5-9\_vs\_7-8       & 0.59422          & 0.65145          & 0.62609             & 0.60664          & 0.6357           & 0.61711          & 0.61383          & \textbf{0.67423} \\
	yeast-0-2-5-6\_vs\_3-7-8-9   & 0.76162          & 0.71633          & 0.76412              & 0.76737          & 0.67056          & \textbf{0.77547} & 0.77521          & 0.67288          \\
	yeast-0-2-5-7-9\_vs\_3-6-8   & \textbf{0.83846} & 0.83431          & 0.83509              & 0.82714          & 0.77885          & 0.83511          & 0.83511          & 0.79958          \\
	yeast-0-5-6-7-9\_vs\_4       & 0.66529          & 0.67508          & 0.70125              & 0.65045          & 0.71723          & 0.66861          & 0.6642           & \textbf{0.72877} \\
	yeast-1-4-5-8\_vs\_7         & 0.50103          & 0.51245          & 0.54385              & 0.52686          & 0.52707          & 0.55142          & 0.55142          & \textbf{0.59496} \\
yeast-1-2-8-9\_vs\_7         & 0.5909           & 0.62314          & 0.64471              & \textbf{0.64714} & 0.62425          & 0.64292          & 0.64292          & 0.64189          \\
	yeast4                       & 0.73483          & 0.66562          & 0.73381              & 0.74519          & 0.68998          & \textbf{0.7604}  & 0.75543          & 0.69003          \\
	yeast5                       & 0.89893          & 0.69963          & 0.87503             & 0.89069          & 0.69937          & \textbf{0.91822} & 0.91698          & 0.65394          \\
	yeast6                       & 0.76903          & 0.68217          & 0.76904              & 0.76317          & 0.60386          & \textbf{0.77044} & 0.76388          & 0.58151          \\ 
\hline
aloi-unsupervised            & 0.53921          & 0.7481           & 0.54976     & 0.54967          & 0.53134          & 0.54941          & 0.54948          & \textbf{0.75685}          \\
annthyroid-unsupervised      & 0.73783          & \textbf{0.907}   & 0.7274               & 0.70221          & 0.75076          & 0.72977          & 0.7293           & 0.88265          \\
	breast-cancer-unsupervised   & 0.98252          & \textbf{0.98564} & 0.9853               & 0.98284          & 0.95324          & 0.98474          & 0.98497          & 0.98215          \\
letter-unsupervised          & 0.62694          & 0.86185          & 0.61467              & 0.52616          & 0.84522          & 0.50224          & 0.56222          & \textbf{0.93071} \\
satellite-unsupervised       & 0.94935          & \textbf{0.97735} & 0.93794              & 0.89599          & 0.89936          & 0.85647          & 0.90062          & 0.9505           \\
	shuttle-unsupervised         & 0.99584          & \textbf{0.9997}  & 0.99683              & 0.99397          & 0.9966           & 0.99397          & 0.99539          & 0.99895          \\
	speech-unsupervised          & 0.47833          & 0.49844          & 0.46786             & 0.47074          & \textbf{0.55401} & 0.47249          & 0.47249          & 0.51043          \\
	pen-global-unsupervised      & 0.94747          & 0.95741          & 0.97291              & 0.86979          & 0.98076          & 0.82344          & 0.90298          & \textbf{0.99562} \\
	pen-local-unsupervised       & 0.77887          & \textbf{0.985}   & 0.5892              & 0.44098          & 0.72426          & 0.43778          & 0.60053          & 0.95085  \\
 \hline    
Best performance & 2 & 7 & 0 & 2 & 1&4 &2& 11\\
 \hline   
	\end{tabular}
	\end{table}

\begin{table}
\caption{Average AUC of various OCC algorithms against the OCCER algorithm with different regression techniques on various domain datasets presented in Table 3. Bold numbers indicate the best performance. }
	\begin{tabular}{|l|l|l|l|l|l|l|l|l|}
\hline
	Dataset                      & IF        & LOF         & OCSVM          & Autoencoder  & OCCER   & OCCER   & OCCER & OCCER
\\    &    &  & & &(ridge) &(lasso) &(elastic) &(RF) 
      \\
\hline
	MF                                   & 0.96915          & 0.89096          & 0.97866    &  0.9414       & 0.98058        & 0.89465        & 0.95099          & \textbf{0.9856}  \\
	COV                                  & 0.83141          & 0.91265          & 0.80444    &  0.76949      & 0.88576        & 0.74736        & 0.76754          & \textbf{0.91432} \\
	DLR                                 & 0.94731          & \textbf{0.98847} & 0.95538    & 0.9364       & 0.98195        & 0.91726        & 0.93566          & 0.98725          \\
	class-level-kc1-defectornot          & \textbf{0.79795} & 0.76268          & 0.70594    &  0.60767      & 0.74634        & 0.61991        & 0.63866          & 0.78619          \\
	kc2                                  & \textbf{0.83985} & 0.63217          & 0.80671    &  0.75406      & 0.79379        & 0.78244        & 0.79177          & 0.82481          \\
	kc1                                  & 0.79241          & 0.6347           & 0.70886    &  0.63417      & 0.73389        & 0.63126        & 0.65868          & \textbf{0.8149}  \\
	cm1                                  & 0.70441          & 0.6617           & 0.63667    &  0.51828      & 0.71357        & 0.54           & 0.55974          & \textbf{0.74785} \\
	datatrieve                           & 0.72892          & 0.69005          & 0.69194    &  0.57184      & 0.62296        & 0.59851        & 0.6172           & \textbf{0.73341} \\
	pc1                                  & 0.69765          & 0.68879          & 0.67564    &  0.59895      & 0.70285        & 0.59778        & 0.61756          & \textbf{0.73167} \\
	class-level-kc1 & \textbf{0.90353} & 0.88423          & 0.86422    &  0.78042      & 0.8163         & 0.77771        & 0.79235          & 0.8919 \\
-defect-count-ranking &  &    &   &    &  &   &     & \\
 \hline      
 Best performance  & 3&1&0&0&0&0&0&6\\
 \hline
	\end{tabular}
	\end{table}
\twocolumn

\begin{table}[]
\caption{Statistical comparison between different pairs of OCC algorithms. Bold numbers indicate the statistically significant difference (p $<$ 0.05) between these two OCC algorithms.} 
\begin{tabular}{|l|l|}
\hline
 Hypothesis         & p value                \\
\hline
Autoencoder vs	OCCER(RF) & \bf{3.2437926776358816E-5} \\
OCSVM vs	 OCCER(RF)   & \bf{0.019943895052829716}  \\
LOF vs	Autoencoder    & \bf{0.031347740640368126}  \\
 IF vs	 OCCER(RF)      & \bf{0.031347740640368126}  \\
 IF vs Autoencoder     & 0.2126935308543134    \\
 LOF vs OCCER(RF)     & 0.2126935308543134    \\
 OCSVM vs Autoencoder & 0.2126935308543134    \\
 LOF vs	OCSVM      & 0.8482363797983206    \\
 IF vs LOF         & 0.8482363797983206    \\
IF vs	OCSVM     & 0.8482363797983206  \\
\hline 
\end{tabular}
\end{table}

\begin{sidewaystable}
\caption{Avrage AUC of various OCC algorithms against OCCER method with different regression techniques on MNIST image dataset.}
	\begin{tabular}{|l|l|l|l|l|l|l|l|l|l|}
\hline
	dataset                      & IF        & LOF         & OCSVM          & Autoencoder  &Autoencoder(I)& OCCER   & OCCER   & OCCER & OCCER
\\    &    &  & & & &(ridge) &(lasso) &(elastic) &(RF)  \\
\hline
mnist-0 & 0.8404  & 0.93351          & 0.85061    &  0.82172      & 0.95441                 & 0.96211          & 0.80828        & 0.84362          & \textbf{0.96545} \\
	mnist-1 & 0.95933 & 0.99122          & 0.97484    &  0.92741      & 0.98311                 & 0.99443          & 0.91213        & 0.95203          & \textbf{0.99591} \\
	mnist-2 & 0.66328 & 0.83744          & 0.672      &  0.64059      & 0.79848                 & 0.83684          & 0.63002        & 0.66314          & \textbf{0.84315} \\
	mnist-3 & 0.65632 & 0.85328          & 0.65381    & 0.73235      & 0.83217                 & 0.81276          & 0.60433        & 0.64505          & \textbf{0.86211} \\
	mnist-4 & 0.62975 & 0.85557          & 0.70193    &  0.59538      & \textbf{0.89449}        & 0.88681          & 0.57024        & 0.64429          & 0.8852           \\
	mnist-5 & 0.52026 & \textbf{0.86928} & 0.49272    &  0.47628      & 0.75357                 & 0.83826          & 0.4712         & 0.51029          & 0.86798          \\
	mnist-6 & 0.74093 & 0.96936          & 0.7386     &  0.65338      & 0.95646                 & \textbf{0.97133} & 0.63303        & 0.69906          & 0.94527          \\
	mnist-7 & 0.7971  & 0.93732          & 0.82775    &  0.72797      & 0.93301                 & 0.90877          & 0.68762        & 0.77317          & \textbf{0.93839} \\
	mnist-8 & 0.64922 & 0.77022          & 0.63571    &  0.62482      & \textbf{0.83047}        & 0.8107           & 0.61917        & 0.64334          & 0.8101           \\
	mnist-9 & 0.71239 & 0.9267           & 0.71086    &  0.6468       & 0.92696                 & \textbf{0.94439} & 0.61743        & 0.68868          & 0.92386       \\  
 \hline      
 Best performance &0 & 1 &0 & 0 & 2 &2&0 & 0 & 5 \\
\hline
	\end{tabular}
	\end{sidewaystable}
\begin{sidewaystable}
	\scriptsize
\caption{AUC of OCCER method with various regression techniques on different domain datasets (Table 3) for different sizes of ensembles. Bold numbers show the best results. CL-KC* is the short form for class-level-kc1-defectornot whereas CL-KC1* is short form for class-level-kc1-defect-count-ranking.}
\begin{tabular}{|l|l|l|l|l|l|l|l|l|l|l|l|l|l|l|l|l|}
\hline
dataset                              & OCCER & OCCER&  OCCER &OCCER & OCCER & OCCER&  OCCER &OCCER & OCCER & OCCER&  OCCER &OCCER & OCCER & OCCER&  OCCER &OCCER    \\                              & (ridge) & (ridge) & (ridge) & (ridge)& (lasso) & (lasso) & (lasso) & (lasso) & (elastic) & (elastic) & (elastic) & (elastic) & (RF) & (RF)    & (RF) & (RF)     \\
          &(25\%) &(50\%)&(75\%) & (100\%)& (25\%) &(50\%)&(75\%) & (100\%)&(25\%) &(50\%)&(75\%) & (100\%) & (25\%) &(50\%)&(75\%) & (100\%) \\
\hline
mf                                   & 0.81275               & 0.72915         & 0.80959                 & 0.98058        & 0.86403               & 0.87841         & 0.89497                 & 0.89465        & 0.86403                 & 0.82846           & 0.85102                   & 0.95099          & 0.81353            & 0.9093           & 0.92949              & \textbf{0.9856}  \\
cov                                  & 0.88756               & 0.89245         & 0.89441                 & 0.88576        & 0.74438               & 0.75445         & 0.74946                 & 0.74736        & 0.74438                 & 0.76094           & 0.75867                   & 0.76754          & 0.91138            & \textbf{0.91806} & 0.91254              & 0.91432          \\
dlr                                  & 0.97361               & 0.97088         & 0.97106                 & 0.98195        & 0.88323               & 0.91032         & 0.91761                 & 0.91726        & 0.88323                 & 0.88321           & 0.9134                    & 0.93566          & 0.98019            & \textbf{0.9885}  & 0.98569              & 0.98725          \\
CL-KC*          & 0.80574               & 0.80219         & 0.79036                 & 0.74634        & 0.5853                & 0.61663         & 0.61796                 & 0.61991        & 0.5853                  & 0.64425           & 0.65478                   & 0.63866          & 0.77922            & 0.81278          & \textbf{0.8154}      & 0.78619          \\
kc2                                  & 0.77763               & 0.79779         & 0.78923                 & 0.79379        & 0.7734                & 0.78226         & 0.77951                 & 0.78244        & 0.7734                  & 0.77638           & 0.78649                   & 0.79177          & 0.80939            & 0.81467          & 0.81941              & \textbf{0.82481} \\
kc1                                  & 0.7656                & 0.77043         & 0.75578                 & 0.73389        & 0.62936               & 0.63398         & 0.63145                 & 0.63126        & 0.62936                 & 0.63409           & 0.64945                   & 0.65868          & 0.79707            & 0.81575          & \textbf{0.81839}     & 0.8149           \\
cm1                                  & 0.62938               & 0.69033         & 0.71871                 & 0.71357        & 0.55644               & 0.54542         & 0.54024                 & 0.54           & 0.55644                 & 0.54294           & 0.56536                   & 0.55974          & 0.69396            & 0.72312          & 0.73804              & \textbf{0.74785} \\
datatreive                           & 0.57859               & 0.56149         & 0.6538                  & 0.62296        & 0.5784                & 0.57582         & 0.59471                 & 0.59851        & 0.5784                  & 0.53139           & 0.56039                   & 0.6172           & 0.64871            & 0.71992          & 0.70694              & \textbf{0.73341} \\
pc1                                  & 0.67077               & 0.64575         & 0.68015                 & 0.70285        & 0.61526               & 0.60874         & 0.5999                  & 0.59778        & 0.61526                 & 0.58243           & 0.59209                   & 0.61756          & 0.70938            & 0.71232          & 0.70353              & \textbf{0.73167} \\
CL-KC1* & 0.85556               & 0.8535          & 0.82941                 & 0.8163         & 0.71203               & 0.77294         & 0.7736                  & 0.77771        & 0.71203                 & 0.76417           & 0.77924                   & 0.79235          & 0.83465            & 0.86493          & 0.8694               & \textbf{0.8919} \\
\hline
\end{tabular}
\end{sidewaystable}

\section{Conclusion}
OCC is a challenging task due to the absence of the outlier class data points in the training dataset.
In this paper, we extensively studied OCCER. OCCER creates $m$ regression models that are used to find outlier scores. As these $m$ regression models are trained only using data belonging to the target class, these models generate more outlier scores for outlier data points. Experiments were conducted  to show the effectiveness of the OCCER, and OCCER performed better than or similar to other OCC methods. OCCER can work well with the latent features created by autoencoders, as shown by the MNIST image data. 

OCCER suggests that regression methods can be used for OCC, and combines the results of  $m$ regression models by averaging them. The other combination techniques will be investigated in the future. The combination of OCCER with other ensemble approaches, such as bagging \cite{Breiman1996} (to create different training datasets),  is another future research direction. In this paper, we investigated the performance of OCCER for image datasets by using the latent features created by autoencoders. Random projections  and principal component analysis are some of the approaches to create new features. We will study the performance of OCCER in the feature space created by random projections  and principal component analysis.


%





\ifCLASSOPTIONcaptionsoff
  \newpage
\fi



%
\bibliography{newoneclass}
\bibliographystyle{IEEEtran}

%





\end{document}